\documentclass[preprint]{elsarticle}
\usepackage{caption,subcaption}
\usepackage{float}
\usepackage{hyperref}
\usepackage{amsmath,amssymb,amsfonts}
\usepackage{xcolor}
\usepackage[normalem]{ulem}
\usepackage{multirow}

\begin{document}

\begin{frontmatter}

\title{CLUE-AI: A Convolutional Three-stream Anomaly Identification Framework for Robot Manipulation}

\author{Dogan Altan\corref{mycorrespondingauthor}}
\cortext[mycorrespondingauthor]{Corresponding author}
\ead{daltan@itu.edu.tr}
\author{Sanem Sariel}
\address{Artificial Intelligence and Robotics Laboratory, Faculty of Computer and Informatics Engineering, Istanbul Technical University, Istanbul, Turkey. }
\address{e-mail: \{daltan,sariel\}@itu.edu.tr}



\begin{abstract}
Robot safety has been a prominent research topic in recent years since robots are more involved in daily tasks. It is crucial to devise the required safety mechanisms to enable service robots to be aware of and react to anomalies (i.e., unexpected deviations from intended outcomes) that arise during the execution of these tasks. Detection and identification of these anomalies is an essential step towards fulfilling these requirements. Although several architectures are proposed for anomaly detection, identification is not yet thoroughly investigated.  This task is challenging since indicators may appear long before anomalies are detected. In this paper, we propose a ConvoLUtional threE-stream Anomaly Identification (CLUE-AI) framework to address this problem. The framework fuses visual, auditory and proprioceptive data streams to identify everyday object manipulation anomalies. A stream of 2D images gathered through an RGB-D camera placed on the head of the robot is processed within a self-attention enabled visual stage to capture visual anomaly indicators. The auditory modality provided by the microphone placed on the robot's lower torso is processed within a designed convolutional neural network (CNN) in the auditory stage. Last, the force applied by the gripper and the gripper state are processed within a CNN to obtain proprioceptive features. These outputs are then combined with a late fusion scheme. Our novel three-stream framework design is analyzed on everyday object manipulation tasks with a Baxter humanoid robot in a semi-structured setting. The results indicate that the framework achieves an f-score of 94\% outperforming the other baselines in classifying anomalies that arise during runtime.
\end{abstract}


\end{frontmatter}


\section{Introduction}

Safe execution is of great importance from the robot ethics perspective \cite{Grinbaum2017,Lin2011}. Ethical robots should operate without any potential damages to humans and their environments. However, anomalies are unavoidable in real world due to various sources of uncertainty. In such cases, at least it is crucial to detect and identify these anomalies. To do so, diagnostic procedures are needed to recognize these cases and recover from them \cite{Ak19whentostop}. The first phase of a diagnostic procedure is being aware of the anomaly, called \textit{anomaly detection}. This is followed by the classification of the anomaly type. This procedure is called \textit{anomaly identification}. After identifying the anomaly, the robot should apply recovery actions to return to the nominal state and achieve the task. This is called \textit{recovery}. Our main motivation in this research is the anomaly identification task where the robot classifies anomaly cases after they are detected. Such a task is vital as being aware of the occurred anomaly type enables the robots to come up with effective recovery plans by associating anomaly contexts with upcoming plans \cite{Karapinar2015}.

\begin{figure}[h]
\captionsetup[subfigure]{}
    \centering
    \begin{subfigure}[t]{0.5\textwidth}
        \centering
        \includegraphics[width=2in]{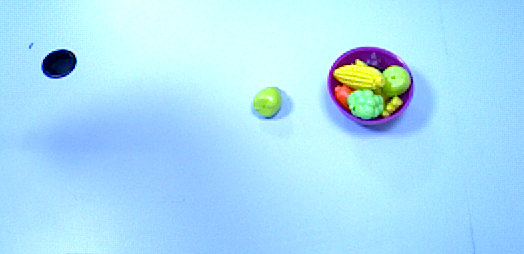}
       \caption{\label{bowl-anomaly-a}}
    \end{subfigure}%
    ~ 
    \begin{subfigure}[t]{0.5\textwidth}
        \centering
        \includegraphics[width=2in]{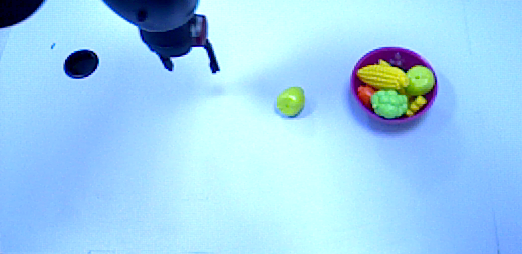}
        \caption{\label{bowl-anomaly-b}}
    \end{subfigure}%
    
     \centering
    \begin{subfigure}[t]{0.5\textwidth}
        \centering
        \includegraphics[width=2in]{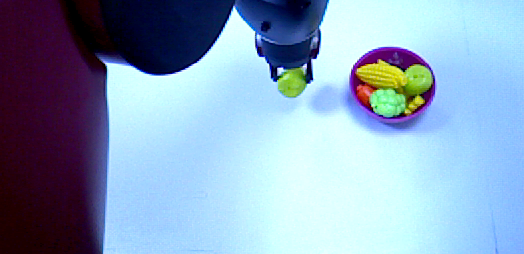}
       \caption{\label{bowl-anomaly-c}}
    \end{subfigure}%
    ~ 
    \begin{subfigure}[t]{0.5\textwidth}
        \centering
        \includegraphics[width=2in]{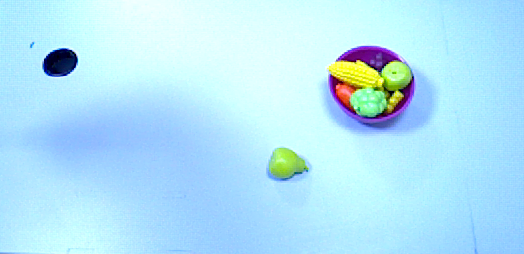}
        \caption{\label{bowl-anomaly-d}}
    \end{subfigure}%
    \caption{\label{bowl-anomaly} An anomalous execution from the viewpoint of a robot with a parallel gripper hand. The task is to place the plastic pear into the container. 
    }
\end{figure}

In this study, we consider robot manipulation anomalies as deviations from rules, specifications, or expectations. Based on this definition, the situations where nature or human interferes with the scene during manipulation are considered anomalies. Anomaly definition also includes violations of expected operational outcomes, failures encountered due to improper computation of grasp, push, place placements, or unsafe executions. Moreover, some anomaly cases may appear although the robot operates safely in a normal situation. A sample case is illustrated in Figure \ref{bowl-anomaly} from the view of the robot. The robot is tasked to place an object (a plastic pear) into a container full of other objects (Figure \ref{bowl-anomaly-a}). First, the robot reaches the pear by its arm and starts executing the plan (Figure \ref{bowl-anomaly-b}). However, since the container is almost full of objects (which is not an anomalous situation at that particular time frame), the plastic pear bounces back and falls onto the table during a safe place-in-container operation (Figure \ref{bowl-anomaly-c} and \ref{bowl-anomaly-d}). The underlying reason of this anomaly is that the robot does not possess the ability to asses the depth of the container and predict that the place operation will fail. At this point, the anomaly identification procedure is expected to ensure that the robot is aware of the fact that a place action is failed due to the situation of the container. This is the key to take necessary precautions for handling future place-in-container tasks on the same container or in a similar case. 

Incorporating sensory readings is the key to an effective anomaly identification procedure. Indeed, a single sensor modality may not be sufficient to accurately identify different cases. Different sensory data may provide a different contribution to the identification process \cite{inceoglu2018comparative}. 
Therefore, gathering information through different sensors is more beneficial. Yet this information should be combined and fused effectively to identify what went wrong during the task operation.

In this paper, we propose a convolutional three-stream anomaly identification (CLUE-AI) framework that processes multimodal data within distinct stages to classify anomaly cases. In our previous work \cite{Altan2021}, we presented a symbol-based anomaly identification framework that adopts an early fusion scheme of multimodal data to identify anomalies. The CLUE-AI framework takes into account visual, auditory and proprioceptive sensor modalities to reveal the anomaly types during execution. Visual data stream is processed taking into account the contribution of each image by enabling a self-attention mechanism. Auditory and proprioceptive data streams are also processed within distinct CNN designs to extract anomaly-related features. A late fusion scheme combines the outputs of these stages to capture anomaly indicators obtained in distinct sensor modality stages. Different from the earlier work \cite{Altan2021}, the CLUE-AI framework does not require any hand-crafted features or domain symbols, and it adopts a late fusion mechanism to fuse attention-enabled visual, auditory and proprioceptive streams that are processed within distinct stages. The CLUE-AI framework is evaluated on real-world scenarios performed by a Baxter robot. A comparison of the framework with other baselines is presented, and the performance of the framework is analyzed for different feature extraction techniques. An ablation study is also presented to analyze the contribution of each sensory modality and the attention mechanism in classifying the anomaly type. The main contributions of this study are three-fold:

\begin{itemize}
    \item We propose a novel convolutional three-stream anomaly identification framework, namely CLUE-AI, that incorporates visual modality together with auditory and proprioceptive modalities to capture anomaly indicators for object-related perceptions. These multimodal data are processed within different stages to identify everyday object manipulation task anomalies.
    \item We deal with cases where anomaly indicators appear long before an anomaly is detected with a self-attention enabled long short-term memory (LSTM) design.
    \item We address the identification of anomaly cases that arise during object manipulation episodes in semi-structured environments. In such environments, task specifications and/or object placements are not fixed or stable, unlike in the case of engineered settings. 
\end{itemize} 

This paper is organized as follows: First, the literature on anomaly identification is summarized. This section is followed by the proposed CLUE-AI framework. Later on, the evaluation of the framework is presented which includes an ablation study and an experimental evaluation of the presented framework. Finally, the paper is concluded with potential future directions.

\section{Literature Review}

Anomaly detection and identification have been widely investigated in the literature \cite{Pettersson2005,Fritz2005} and several taxonomies of failures that could occur in task environments are presented \cite{Carlson2004,Karapinar2012,Lopez2017}. This section presents a summary of the related work on the anomaly identification literature.

Anomaly identification can be achieved by using hypothesis-based methods by either maintaining cost-attached hypotheses in a hypothesis pool and analyzing inconsistencies among them \cite{Gspandl2012} or comparing the differences between the theory and the model of the world \cite{Steinbauer2009}. A co-operative diagnostic method may also be presented for a multi-robot domain to diagnose failures where robots help each other to do so \cite{Morais2015}.

Clustering-based methods are also investigated to identify failure cases. In one study \cite{Abid2015}, a multi-level sensor fusion method is investigated to detect and identify abnormal cases by clustering the outputs of the sensors. In another clustering-based method, a global fuzzy c-means clustering algorithm is used to maintain clusters to identify failures \cite{Schleyer2011}. Combining unsupervised learning with supervised learning is also studied in the literature. In a clustering-based method \cite{Biswas2016}, outliers in the data are first detected with unsupervised learning techniques, which are followed by differentiating between special modes and anomalies with supervised techniques. 

In the literature, particle filter-based (PF) \cite{Verma2004}, Kalman filter-based \cite{Rigatos2009} and hidden Markov model-based (HMM)  \cite{Altan2014,Altan2016Empirical,inceoglu2018comparative,inceoglu2018failure} methods are widely used for handling anomalies. Various types of HMMs are used for detecting identifying failures that occur during assistance tasks \cite{Park2018HMM}. In one study \cite{Luo2018}, HMMs are studied with gradient analysis to identify anomaly cases. Bayesian filters are also studied to analyze failures \cite{Di2013}. A hierarchy for HMMs and PFs is proposed to isolate failure cases on a mobile robot setting \cite{Altan2014,altan2014hierarchical,Altan2016Empirical}. In another work \cite{Long2018risk}, a probabilistic method is proposed to predict failure cases for humanoid robots in hazardous environments by associating risks with related actions. 

Deep learning based methods are also applied to handle anomaly cases. An autoencoder-based method that uses stacked denoising autoencoder (SDA) is proposed \cite{Lu2017fault} to identify the health state of rotary machines. Transfer learning (TL) is also studied to diagnose industrial failures. A deep transfer learning (DTL) method is proposed to handle motor bearing failures \cite{Wen2019}. The model of task execution can be constructed with convolutional neural networks (CNNs) to identify anomalies \cite{Bowkett2018}. In another work, anomalies are detected with a multimodal sensor fusion based deep neural network design \cite{Inceoglu21FINO}. Another autoencoder based method adopts variational LSTM autoencoder to detect anomalies in robot assisted feeding tasks \cite{park2018multimodal}. In yet another study, multilayer perceptrons (MLPs) combine the temporal dependencies in multimodal data captured by HMMs, and the convolutional features are extracted from visual data by VGG16 to classify anomaly types in the domain of human-robot interaction in human feeding scenarios \cite{Park2017multimodal}. Different from this study, we address everyday object manipulation anomalies that arise due to uncertainties in perception and/or execution failures. Moreover, instead of extracting temporal features with HMMs and temporal pyramid pooling \cite{madry2014st} to capture unexpected trends in the data for a fixed amount of time steps, we employ self-attention enabled LSTMs to capture anomaly indicators that may be observed long before the anomaly occurrences. In our previous work \cite{Altan2021}, we present a symbolic-level anomaly identification method that processes the outputs of a visual scene modeling system \cite{Inceoglu2018}, proprioceptive sensors and auditory data to identify anomalies with preprocessed hand-crafted features. In this study, we extend it by presenting a three stream anomaly identification framework that extracts low-level features from 2D images directly without considering high-level symbolic domain symbols which does not require any hand-crafted feature engineering effort. Furthermore, we deal with a more extensive set of anomaly cases for a service robot performing everyday tabletop scenarios. 

\section{Convolutional Three-stream Anomaly Identification (CLUE-AI) Framework}

The CLUE-AI framework consists of three steps of processing three different sensory streams. The visual data (2D images) collected from an RGB-D camera (ASUS Xtion RDB-D Camera) mounted on the robot's head is processed in the first stage. The second stage analyzes auditory data obtained by a microphone (PSEye microphone) mounted on the Baxter robot's lower torso. The final stage deals with gripper-related data (i.e., the position of the gripper (its openness), the force applied by the gripper to the object at hand). The following subsections elaborate on the procedures that take place on these aforementioned streams. 

\begin{figure}[h]
\begin{center}
     {\includegraphics[scale=0.45]{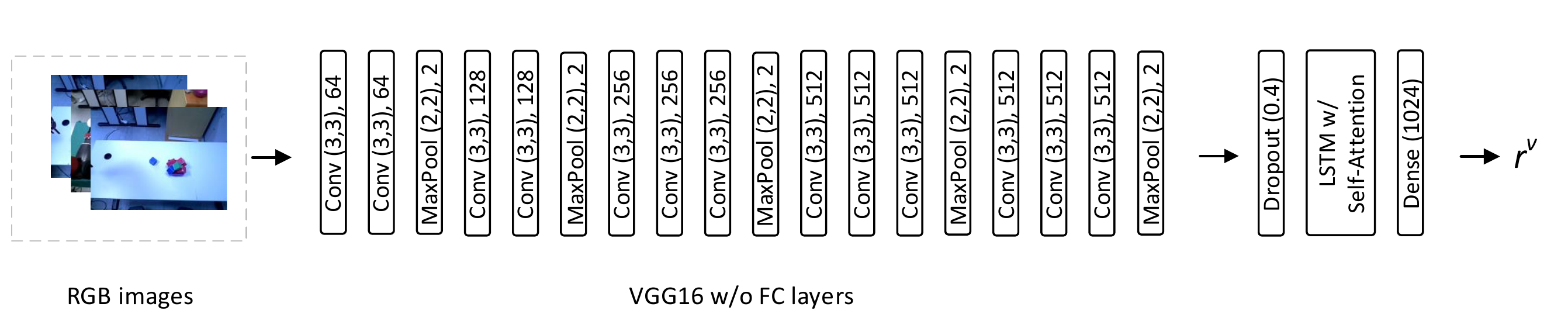}}
\caption{\label{fig:clue-ai-visual}The architecture for processing visual stream.}
\end{center}
\end{figure}

\subsection{Visual stream}

Sequential RGB frames are retrieved from videos obtained from the head camera in this research, and these frames are sampled at a fixed frequency of 0.125 Hz. Using the timestamps of the frames, this sampling method provides a frame sequence that is sorted in ascending order. To capture temporal dependencies among anomaly indicators, convolutional visual features are extracted from each sampled image, and the sequences of these features are fed into the LSTM layers while retaining the order of the sampled frames. These attributes are retrieved from the consecutive 2D images collected by the RGB-D camera mounted on the robot's head using a pre-trained convolutional neural network (CNN) structure. Images are cropped and scaled ($\mathbb{R}^{224x224}$) before the feature extraction procedure. The altered images are sent into a pre-trained CNN structure after this transformation phase to detect the crucial parts of them. The following CNN architectures are utilized to extract features from consecutive 2D images in order to achieve this goal:

\begin{itemize}
    \item \textbf{Residual Net (ResNet)} \cite{He2016}:
    ResNets are neural networks that make training deeper neural networks easier. For deeper networks, they are not only easier to optimize, but they also provide high accuracy. The basic principle underlying this architecture is that instead of learning functions without references, the layers are treated as residual functions that reference the layer inputs. ResNets enables learning with up to 152 layers.
    
    \item \textbf{AlexNet} \cite{Krizhevsky2012}: 
    AlexNet is a CNN architecture with five convolutional layers and three fully-connected layers. A max pooling operation is performed after the first, second, and fifth convolution layers. On the ImageNet dataset \cite{deng2009imagenet}, the design performs high accuracy.

    \item \textbf{VGG} \cite{Simonyan2014}: 
    The principle behind this structure is to use an architecture with  small (3x3) convolution filters to evaluate networks with increasing depths. It typically has 16 to 19 layers. The VGG variation with 16 layers (VGG16) is used in this research. VGG16 has thirteen convolutional layers and three fully-connected layers. Each convolution layer is followed by ReLU, and after the second, fourth, seventh, tenth, and thirteenth convolution layers, a max pooling operation is performed.
    
\end{itemize}

In this research, pre-trained implementations of these vision models on ImageNet \cite{Russakovsky2015} are used. These models are trained on a variety of images, including a variety of objects. The final layers of these pre-trained CNNs, which contain fully-connected layers, are not employed because they are only used for extracting relevant features. $f_t^v$ is the symbol for each feature vector at time $t$. Following the extraction of relevant visual features ($f^v$) related to anomaly indicators, a dropout function with a probability of 0.4 is used. The generated features are then fed into LSTM cells, which are used to learn anomaly patterns. The LSTM result is then sent into an attention layer to assess the importance of the images in the sequence that contribute to the anomaly decision. The LSTM layers in this work adopt scaled dot-product self-attention \cite{Vaswani2017attention} to come up with attention values associated to the images in the sequence. After calculating self-attention scores, they are fed into a dense layer. The output vector is then created, which includes the concatenation of the attention output with the LSTM outputs ($r^v \in \mathbb{R}^{1024}$) and is fused with the auditory and proprioceptive modality outputs.

The steps of the proposed CLUE-AI framework while processing visual modality are depicted in Figure \ref{fig:clue-ai-visual}. To handle sequential RGB images, the proposed visual processing technique involves two phases: extraction of features and learning of these extracted features. Inputs are accepted in the form of sequential 2D RGB images. Green layers in the feature extraction stage represent convolution layers, whereas blue layers represent pooling procedures. Convolution layers are labeled with the kernel sizes and channels that correspond to them. Without its classification layers, which contain fully connected layers, these sequential layers form the VGG network. A dropout layer follows this structure, which is followed by LSTM layers that learn the visual anomaly indicators. The LSTM outputs are then passed into an attention layer, which generates attention values based on the image positions in the sequence. Note that after each convolution layer, a rectified linear unit (ReLU) is utilized as an activation function; these are omitted in the figure for brevity.

\subsection{Auditory stage} 

The audio data collected by the microphone mounted on the robot's torso is processed in two steps. The initial phase involves extracting features from the collected audio data. Mel-frequency Cepstral Coefficients (MFCC) characteristics are used to do so. For this purpose, the library Librosa \cite{Mcfee2015} is utilized. $f_t^a$ represents the extracted auditory features of an observation collected at time step $t$. In the second phase, these extracted MFCC features ($f^a$) are sent into a CNN block. The block is made up of four convolution layers that are stacked one on top of the other. As an activation function, each convolution layer is followed by a ReLU unit. Max-pooling is applied after these procedures, and the resulting vector is fed into a dense layer, which outputs the resulting auditory feature vector ($r^a \in \mathbb{R}^{64}$).

\begin{figure}[h!]
\begin{center}
     {\includegraphics[scale=0.8]{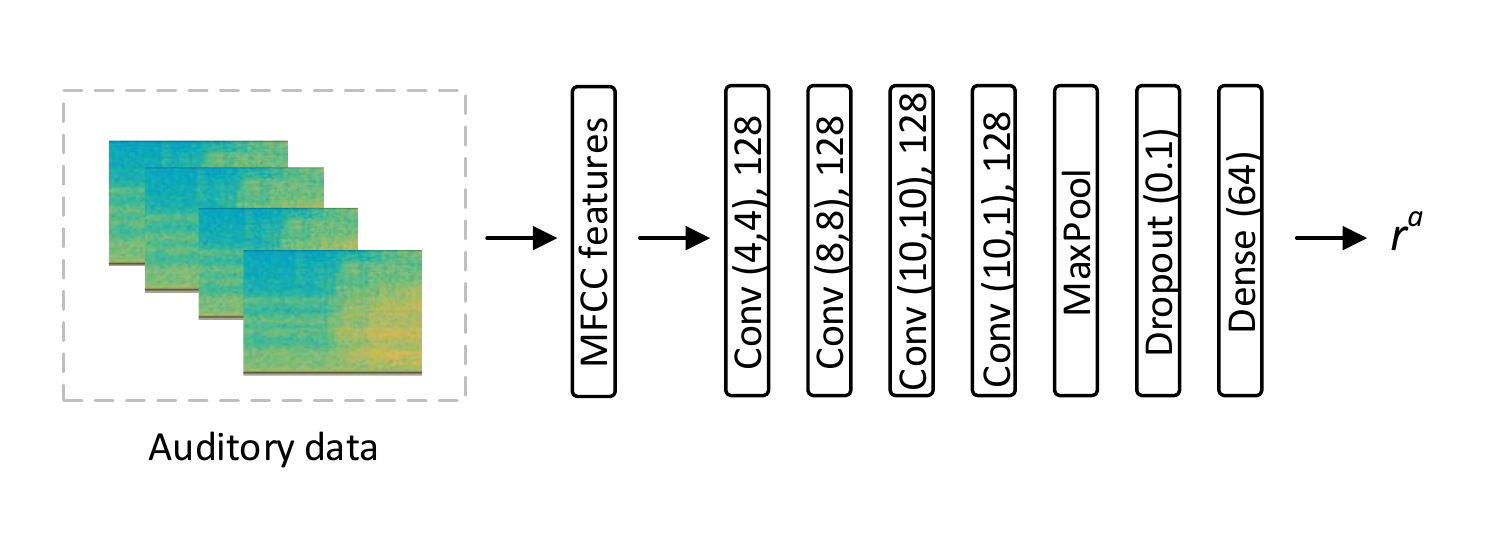}}
\caption{\label{fig:clue-ai-auditory}The architecture for processing auditory stream. }
\end{center}
\end{figure}

The contents of the proposed framework's auditory stage, which analyses audio data, are shown in Figure \ref{fig:clue-ai-auditory}. The audio data is processed in two steps, as shown in the figure. The audio data's MFCC features are extracted first, and then they're processed in convolution layers. The convolution layers' outputs are fed into a dense layer, which creates a vector that includes processed audio features.

\subsection{Proprioceptive stream}

\begin{figure}[h]
\begin{center}
     {\includegraphics[scale=0.8]{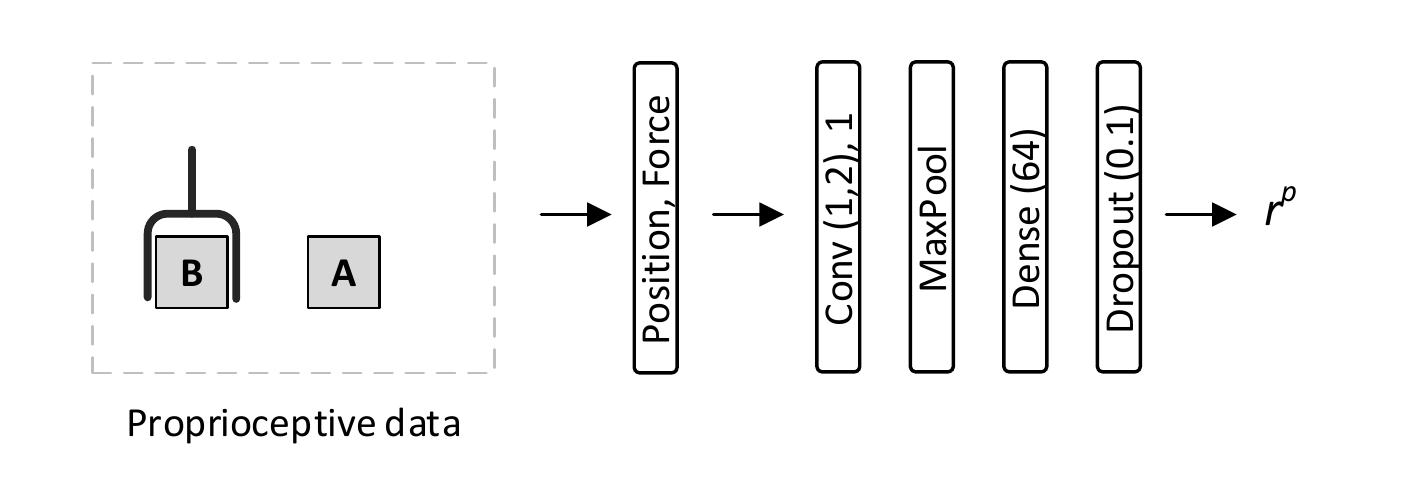}}
\caption{\label{fig:clue-ai-proprioceptive}The architecture for processing proprioceptive stream.}
\end{center}
\end{figure}

The gripper on the robot's arm is used to manipulate objects in the environment. With the gripper, continuous proprioceptive observations are gathered. The gripper's openness and the force applied by the gripper to the object in hand are considered features ($f^p$) in this study. Figure \ref{fig:clue-ai-proprioceptive} depicts how these features are processed within a tiny CNN. This CNN's output is fed into a max pooling layer, which is then followed by a dense and dropout layer. As a result, the final vector for the gripper data are generated ($r^p \in \mathbb{R}^{64}$).

\subsection{Late fusion}

\begin{figure}[h]
\begin{center}
     {\includegraphics[scale=0.8]{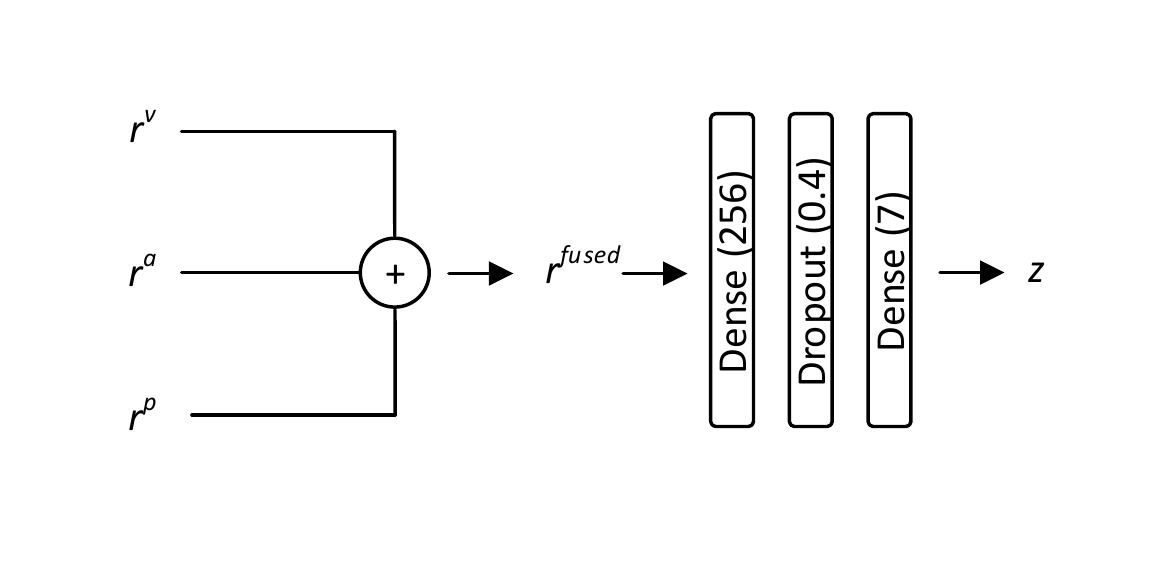}}
\caption{\label{fig:clue-ai-concat}The late fusion scheme of the CLUE-AI framework.}
\end{center}
\end{figure}

The explained feature vectors make up the aforementioned visual, auditory, and proprioceptive streams. These collected visual, auditory, and proprioceptive features are concatenated to a single fused vector ($r^{fused} \in \mathbb{R}^{1152}$) using a late fusion technique. Following that, the fusion's resultant vector is fed into a dense layer, which is then followed by a dropout layer with a probability of 0.4. The resulting vector is then passed into a softmax function and another dense layer. Based on these findings, the most likely anomaly type is picked for the anomaly's identification. The late fusion method is depicted in Figure \ref{fig:clue-ai-concat}.

\section{Experimental Analysis}

The CLUE-AI framework is evaluated in real-world experiments using the Baxter robot performing everyday object manipulation scenarios. This section presents the experimental setup and the results.

\begin{figure}[h]
\includegraphics[width=8cm]{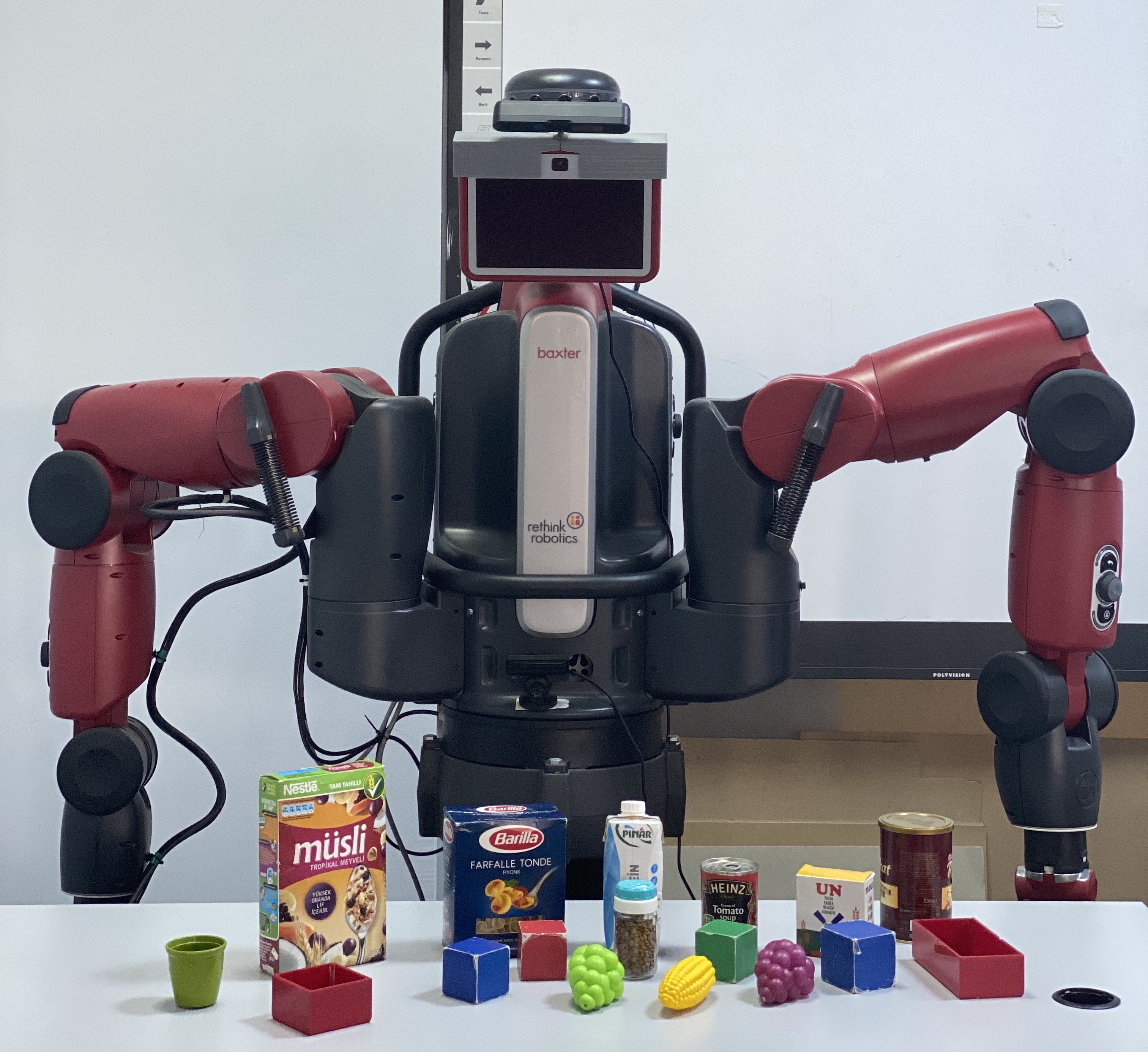}
\centering
\caption{Baxter robot and the objects that are used in the experiments.}
\label{baxter}
\end{figure}

\subsection{Experimental Setup}

In this section, the CLUE-AI framework is evaluated in real-world experiments using the Baxter robot performing everyday object manipulation scenarios. Experiments based on the aforementioned anomaly situations are carried out with an RGB-D camera attached on Baxter's head and a microphone mounted on its body (Figure \ref{baxter}). The robot interacts with objects to perform tasks by executing its actions in everyday object manipulation tasks. We investigate the following types of everyday object manipulation anomalies in this study:

\paragraph{Location Change Anomaly (LOC)} 
This anomaly type refers to situations in which object states (particularly their locations) are changed without the robot's knowledge, yet the objects remain in the scene. Another robot or a human may perform this change.

\paragraph{Object Disappearance Anomaly (DIS)} 
    This anomaly type refers to situations where one or more objects with which the robot interacts are taken out of the environment or from the robot's field of view. Another robot or human in the environment may be responsible for this disappearance.
    
\paragraph{Earlier unstable action anomaly (EUA)} 
    This anomaly occurs when the robot is assigned the task of constructing a tower out of a collection of objects. The tower, however, collapses during construction as a result of an earlier unstable action that results in a misplaced or inappropriately placed sub-tower.
    
\paragraph{Overturning Anomaly (OTA)} 
    To avoid encountering anomaly cases, a robot should select the appropriate push point whenever it needs to push an object. Otherwise, the object may collapse, fall, or shift its orientation. This anomaly type corresponds to such cases.
    
\paragraph{Spilled object anomaly (SPC)} 
    When the robot is tasked to pour objects or liquids into a container, the destination container may be empty, partially full, or full. However, the robot may not recognize the destination container's fullness, causing the destination container to overflow and spilling the contents of the first container onto the table.
    
\paragraph{Full container anomaly (FCA)} 
    Unlike the previously described anomaly type (SPC), this anomaly type refers to situations in which an object is placed or stacked into an already full container. As a result, the object falls as a result of the robot's placement of the object into the container.

The robot can perform five actions in this study. To move its arm to an object location, the robot executes \textit{move-to-object}. To locate its arm to a destination point, it executes \textit{move-to-location}. \textit{pick} is used to grab a target object with the gripper, and \textit{place} is used to place the object to its destination. It can also use the \textit{push} action to move a target object along a specified dimension and distance with its gripper.

Experiments on seven distinct classes involving six anomaly cases and the safe cases are conducted to evaluate the presented framework (SAFE). Safe scenarios are also included in the evaluation to identify and deal with any potential false alarms that may arise during the anomaly detection procedure. Experiments are carried out on a server with the following specs: Intel Core i7-7700K 4.20GHz CPU, 32 GB RAM, and an NVIDIA Quadro P6000 GPU with 24 GB memory. The proposed framework is evaluated using 249 real-world anomaly scenarios collected by the Baxter robot (68 SAFE, 22 LOC, 41 DIS, 33 EUA, 18 OTA, 43 SPC, 24 FCA). When updating the network, an adaptive weighting approach is used to deal with the data set's class imbalance problem, where anomaly classes with the lowest total number of instances are valued more. For each class, the data set is randomly partitioned into the train (80\%) and test (20\%) sets. The following are the parameters are set: For each training epoch, the Adam algorithm  \cite{Kingma2014} is used as an optimizer and cross-entropy loss to calculate the loss. The models are trained for 40 epochs which is set empirically considering the average number of iterations that is sufficient with a learning rate of $10^{-4}$ ($\eta$). The hidden size is 512, and the LSTM structure is made up of one layer. The results are reported based on the average scores on the test set for ten random seeds.

\subsection{Experimental Evaluation}

The CLUE-AI framework is evaluated on a number of aspects. The impact of various feature extraction techniques is first analyzed. After that, a set of analyses such as comparative analysis, ablation analysis, noisy data analysis, and auditory analysis are performed.

\subsubsection{Feature extraction analysis}

The CLUE-AI framework is tested with four different feature extraction settings employing the following CNNs: ResNet18, ResNet101, AlexNet, and VGG16. It's important to note that ResNet18 and ResNet101 are two ResNet variants with different numbers of layers in the corresponding CNN structure.

\begin{figure}[h]
\centering
    \begin{subfigure}[t]{0.45\textwidth}
         \hfill
         {\includegraphics[scale=0.4]{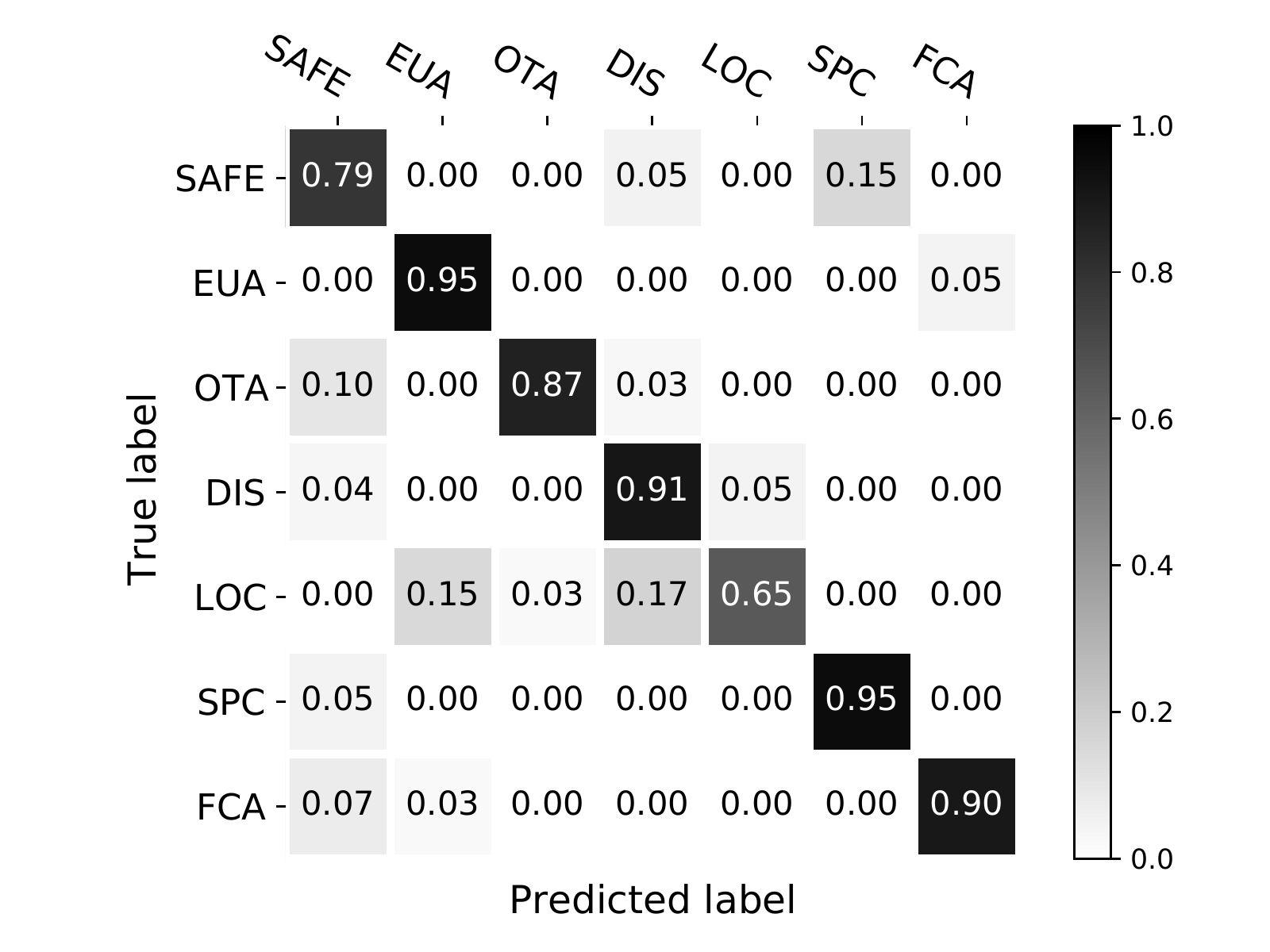}}
\caption{\label{res18-clue}ResNet18 as a feature extractor.}
    \end{subfigure}%
    \begin{subfigure}[t]{0.45\textwidth}
         \hfill
        {\includegraphics[scale=0.4]{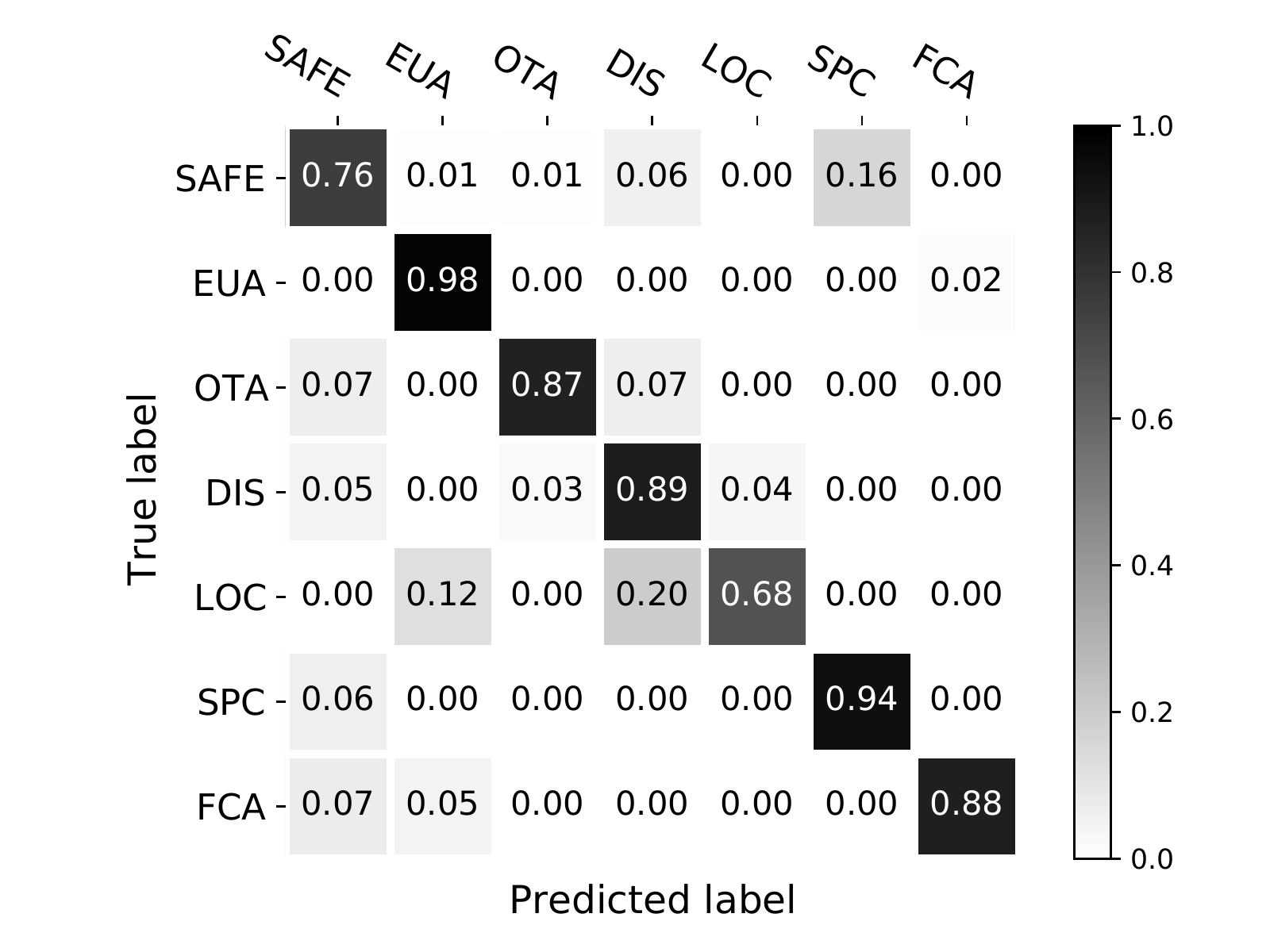}}
\caption{\label{res101-clue}ResNet101 as a feature extractor.}
    \end{subfigure}%

    \centering
    \begin{subfigure}[t]{0.45\textwidth}
         \hfill
         {\includegraphics[scale=0.4]{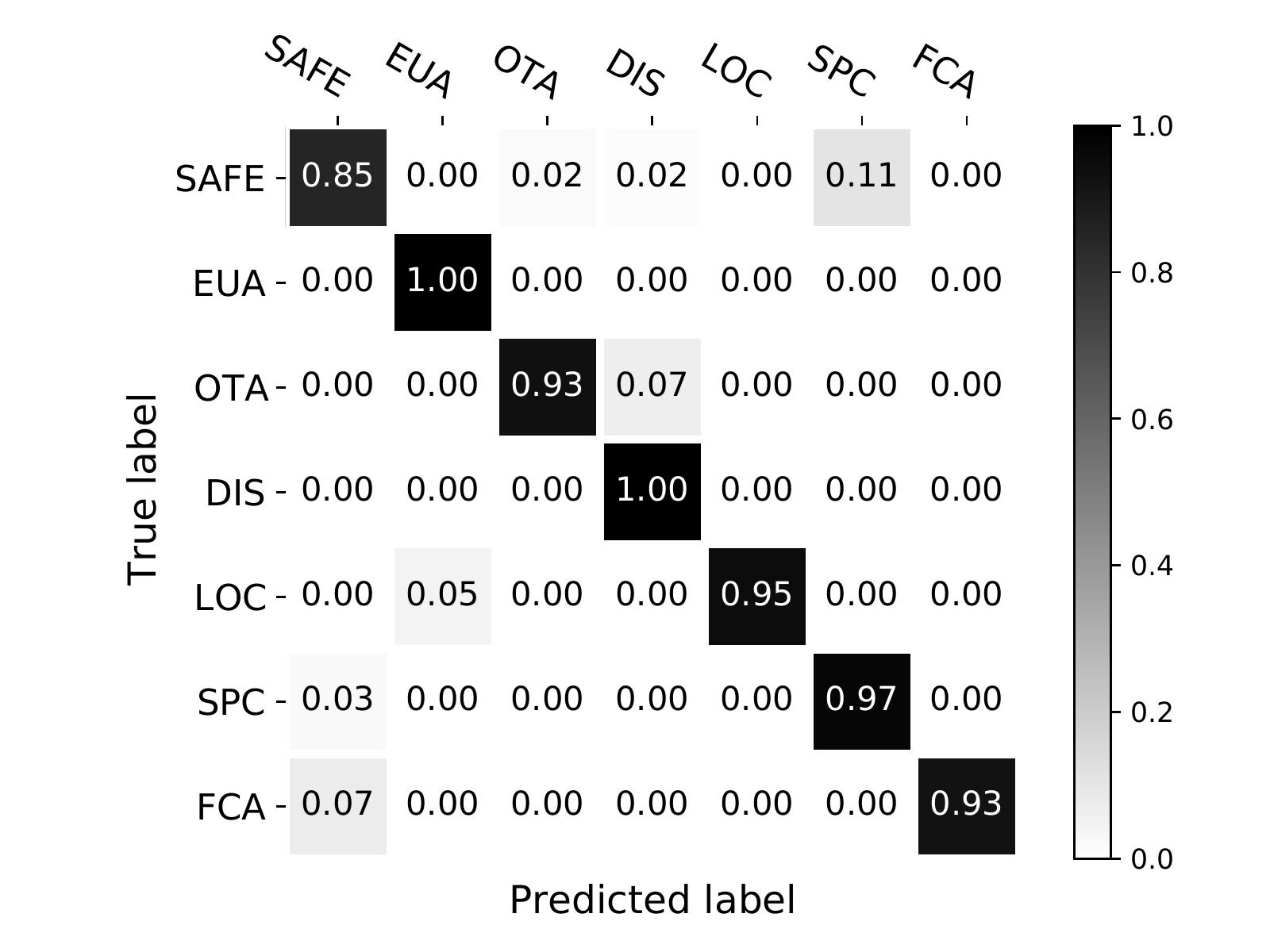}}
\caption{\label{alex-clue}AlexNet as a feature extractor.}
    \end{subfigure}%
    \begin{subfigure}[t]{0.45\textwidth}
         \hfill
        {\includegraphics[scale=0.4]{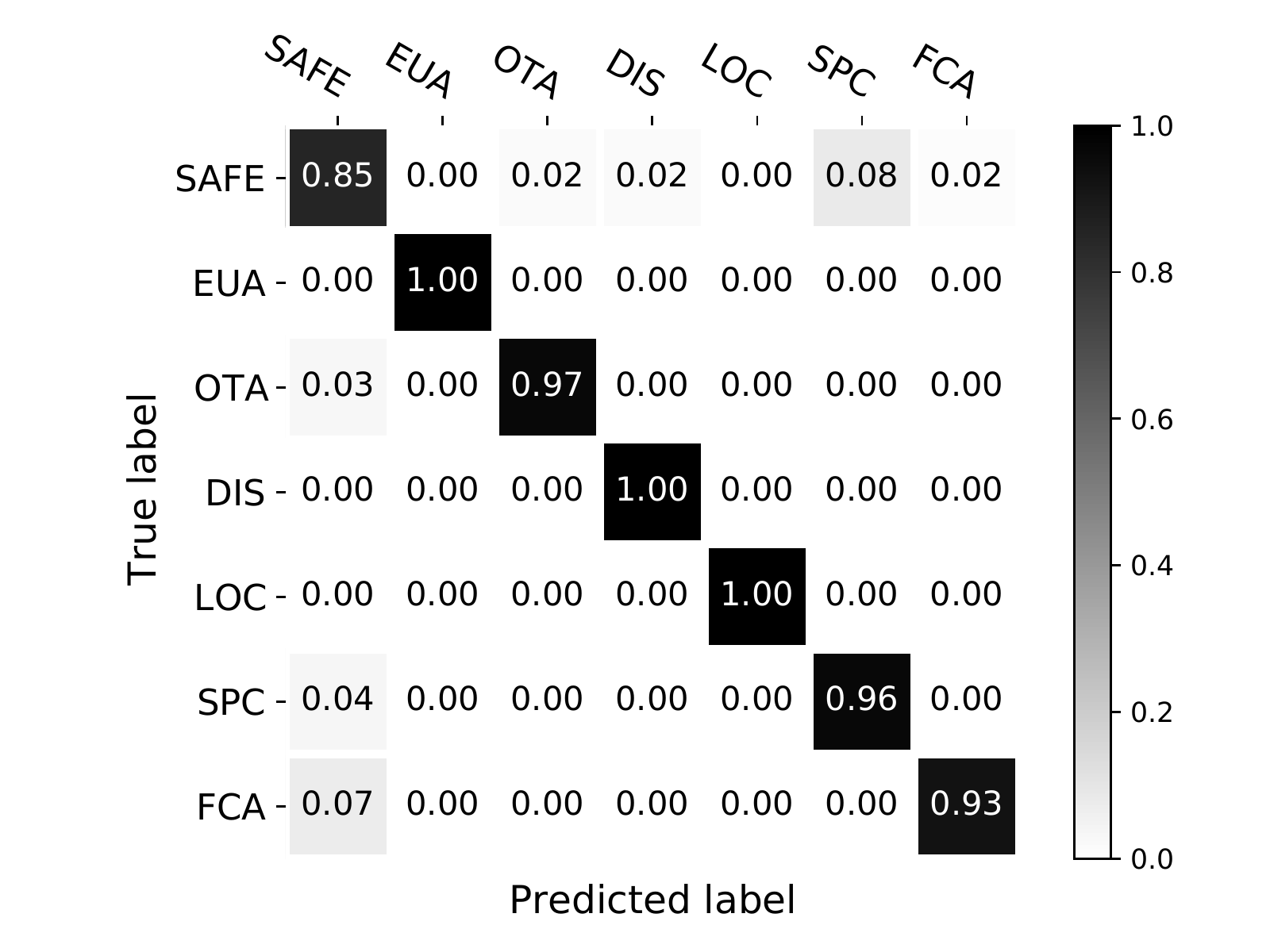}}
\caption{\label{vgg-clue}VGG16 as a feature extractor.}
    \end{subfigure}%
    
    \caption{\label{conf-mat-clue}Normalized confusion matrices of different feature extraction techniques for the visual stream.}
\end{figure}

The normalized confusion matrices for different CNNs that extract visual features are shown in Figures \ref{res18-clue}-\ref{vgg-clue}. Given the results in Figure \ref{res18-clue}, one might conclude that ResNet18 is unable to extract visual features for some classes. The average pooling (AP) layers of ResNets (with a square kernel size of seven) that are placed before the excluded dense layers are the main cause of this situation. As a result, the size of the extracted features shrinks. In this study, only the dense layers of these models are excluded. Since anomaly indicators cannot be distinguished in such a situation, ambiguity in anomaly classes arises. For example, in a ResNet18-based setting, SAFE and LOC are frequently confused classes. LOC (the anomaly cases where an object's location is changed by an external agent) is confused with DIS, as shown in Figure \ref{res18-clue} (the anomaly cases where an object is taken out of the scene by a human). That is, the object's change of location is mistaken for its disappearance. The ResNet101-based setting and the ResNet18-based setting produce similar results. For the anomaly identification task, however, AlexNet and VGG16-based settings provide more accurate classification results (Figures \ref{alex-clue}-\ref{vgg-clue}).

The minimum classification score for the AlexNet structure as the feature extractor belongs to class SAFE among the classes with a value of 0.85. When VGG16 is used to extract relevant information about anomaly symptoms, the SAFE class has a minimum score of 0.85, indicating that no anomaly has occurred. The interpretation of this situation is that a small amount of chickpeas may be spilled on the table, being occluded by the container on some occasions. 

The performance analysis of the presented feature extraction methods in terms of precision, recall, and f-score metrics is summarized in Table \ref{perf-analysis-feature}. Both variations of the ResNet settings with and without average pooling (AP) layers placed last are taken into account in this analysis. In comparison to the setting with AP layers, the ResNet18 setting without AP layers produces better results. The AP layers (with a square kernel size of seven) of ResNets that are placed before the excluded dense layers are the main cause of this situation. As a result of the usage of these AP layers, the size of the extracted features reduces.
Despite the fact that the ResNet-based feature extraction method provides better performances without AP layers, the AlexNet-based feature extraction technique has an f-score of 93.88\%. As shown in the table, VGG16 is capable of extracting relevant visual features, outperforming the other ResNet-based settings and slightly outperforming the AlexNet-based setting with an f-score of 94.34\%.

\begin{table}[h]
\centering
\caption{Performance evaluation of different feature extraction methods.}
\begin{tabular}{c|ccc}
\hline
\multicolumn{1}{l}{}   &   {{Precision}}                              & {{Recall}} & {{F-score}}\\
\multicolumn{1}{l}{}      & {{($\mu \pm \sigma$)}}                          & {{($\mu \pm \sigma$)}} & {{($\mu \pm \sigma$)}} \\
 \hline
\multicolumn{1}{c}{{{ResNet18 w/ AP}}}    &  89.32  $\pm$ 2.17 & 87.60 $\pm$ 2.39   & 87.16 $\pm$ 2.75\\ 
\multicolumn{1}{c}{{{ResNet18 w/o AP}}}    &  92.65  $\pm$ 2.54  & 91.52 $\pm$ 2.82   & 91.43 $\pm$ 2.85\\ 
\multicolumn{1}{c}{\textit{{ ResNet101 w/ AP}}}    &   89.63  $\pm$ 4.50  & 88.04 $\pm$ 4.48  & 87.76 $\pm$ 4.49\\ 
\multicolumn{1}{c}{\textit{{ResNet101 w/o AP}}}    &  91.86   $\pm$ 3.80  & 90.43 $\pm$ 4.03   & 90.33 $\pm$ 4.15\\ 
\multicolumn{1}{c}{{{AlexNet}}}    &  94.63   $\pm$ 2.57  & 93.91 $\pm$ 2.71   & 93.88 $\pm$ 2.76\\ 
\multicolumn{1}{c}{{{VGG16}}}    &  94.90  $\pm$   2.23    &      94.34  $\pm$   2.21      &    \textbf{94.34  $\pm$  2.26}\\\hline
\end{tabular}
\label{perf-analysis-feature}
\end{table}

The elapsed time for training the anomaly models with features extracted by different CNNs is shown in Table  \ref{time-feature}. Each column corresponds to the elapsed times for training and testing in seconds, and each row contains the results of a different feature extraction method (sec). The average and standard deviation of the execution times are reported for ten executions. As can be seen from the results, AlexNet takes the least amount of time to train and test models. The most time (approximately 19 seconds) is required to train for a single epoch using a ResNet101 without average pooling layers and feeding the extracted features into LSTMs.

\begin{table}[h]
\centering
\caption{Time elapsed (sec) during training and testing with different feature extraction techniques.}
\begin{tabular}{ccc}
\hline
                   & {\begin{tabular}[c]{@{}c@{}}Train\\ ($\mu \pm \sigma$)\end{tabular}} & {\begin{tabular}[c]{@{}c@{}}Test\\ ($\mu \pm \sigma$)\end{tabular}} \\ \hline
{CLUE-AI w/ ResNet18 w/ AP}  &   7.43 $\pm$  0.05   &  0.33 $\pm$  0.01    \\ 
{CLUE-AI w/ ResNet18 w/o AP}  &   10.39 $\pm$  0.04   &  1.12 $\pm$  0.01    \\  
{CLUE-AI w/ ResNet101 w/ AP} &  11.35 $\pm$  0.07 &      1.22 $\pm$  0.03      \\  
{CLUE-AI w/ ResNet101 w/o AP} &  18.89 $\pm$ 0.08  &   2.38    $\pm$    0.04    \\  
{CLUE-AI w/ AlexNet}   &  \textbf{7.12 $\pm$  0.05} &   \textbf{0.21 $\pm$  0.01 }   \\  
{CLUE-AI w/ VGG16}   & 11.77 $\pm$  0.07 &  1.20 $\pm$  0.03           \\  \hline  
\end{tabular}
\label{time-feature}
\end{table}

\subsubsection{Overall performance analysis}

The proposed CLUE-AI framework is compared to various methods for the anomaly identification task in this experimental analysis. These methods include hidden Markov models (HMMs) and recurrent neural networks (RNNs). For the RNN-based method, LSTMs are replaced with the corresponding structures in the CLUE-AI framework. To extract relevant features from the input images for the HMM-based method, incremental principal components analysis (IPCA) is used. These features are then fed into HMMs to classify the anomaly type. For each anomaly type, an HMM model with binary latent states (safe and the corresponding anomaly type) is trained. The observation history, including images, is fed into each trained model in the event of an anomaly. As a result, the scores from each anomaly model that corresponds to the likelihood of that anomaly model are generated. The anomaly type is then determined by combining these likelihoods with the results of the auditory stage. The weighted average scores are reported in Table \ref{clue-ai-perf-test} after each method is run ten times with different seeds.

\begin{table}[h]
\centering
\caption{Comparative performance analysis of different methods.}
\begin{tabular}{cccc}
  \hline
\multicolumn{1}{l}{}   &   {{Precision}}                              & {{Recall}} & {{F-score}}\\
\multicolumn{1}{l}{}      & {{($\mu \pm \sigma$)}}                          & {{($\mu \pm \sigma$)}} & {{($\mu \pm \sigma$)}} \\
 \hline
\multicolumn{1}{c}{{{HMM}}}   & 74.05 $\pm$ 4.31    & 71.63 $\pm$ 5.73  & 70.75 $\pm$ 6.08      \\ 
\multicolumn{1}{c}{{{Violet-LSTM \cite{Altan2021}}}}    &  96.20   $\pm$ 3.77  & 89.32 $\pm$ 6.82  & 92.14 $\pm$5.99\\ 
\multicolumn{1}{c}{{{CLUE-AI w/ RNN}}}    & 93.51  $\pm$  3.83 & 92.39 $\pm$ 4.48   & 92.34 $\pm$ 4.51          \\ 
\multicolumn{1}{c}{{{CLUE-AI w/ LSTM}}}    &  94.90  $\pm$   2.23    &      94.34  $\pm$   2.21      &    \textbf{94.34  $\pm$  2.26}\\ \hline

\end{tabular}
\label{clue-ai-perf-test}
\end{table}

In terms of all criterion, VGG16 as the feature extractor with LSTMs (CLUE-AI w/ VGG16-LSTM) outperforms the other deep learning-based methods and HMMs with IPCA as the feature extractor (IPCA-HMM). VGG16 performs an f-score of 92.34 percent when the extracted features are processed with RNNs. When VGG16 and LSTMs are combined in the CLUE-AI framework, they outperform the other methods, with an f-score of 94.34\% in classifying anomaly types.

When compared to the results obtained with the anomaly identification algorithm in \cite{Altan2021} that uses symbols as visual features, it can be concluded that the CLUE-AI framework with VGG16-LSTM outperforms the symbol-based anomaly identification algorithm with an f-score of 94\%. In comparison to the symbol-based anomaly identification algorithm, the CLUE-AI framework is evaluated on a larger set of anomaly scenarios (i.e., a greater number of anomaly classes). Moreover, a challenging set of scenarios for a scene interpretation system is included in this extended set of anomaly scenarios (i.e., cluttered environments or containers full of objects).

\subsubsection{Ablation study}

The CLUE-AI framework proposed in this paper processes data from various sensory modalities. To identify the anomaly types, a multimodality analysis is used to examine the contribution of each processed sensory modality and the attention mechanism.

The precision, recall, and f-score metrics for different sensory modality settings of the CLUE-AI framework are presented in Table \ref{modality-clue}. Furthermore, one column in the table is set aside to indicate whether the attention mechanism is used in the given situation. Each column in the table represents the scores for the corresponding metric, and each row corresponds to a different setting. This setting yields an f-score of 88.70\% when only visual modality with self-attention is considered for classification. When the gripper-related data is combined with the visual modality, the f-score improves slightly (88.72\%).

Combining visual and auditory modalities with self-attention leads to improved performance of an f-score of 93.51\%. Incorporating the auditory modality allows the robot to more effectively distinguish anomaly cases in which sound is a discriminating indicator (for example, when an object falls down as it is pushed, the robot perceives sound as observation; on the other hand, it does not perceive sound when an external agent changes the object's location). As can be seen in the table's last two rows, combining the attention mechanism with the three modality stream results in a 2\% increase in the f-score.

\begin{table}[h]
\centering
\caption{Sensory modality and attention analysis.\label{modality-clue}}
\begin{tabular}{ccccccc}
\hline
\multicolumn{3}{c}{{Modality}} & \multirow{2}{*}{{Attention}} & \multirow{2}{*}{{Precision}} & \multirow{2}{*}{{Recall} }& \multirow{2}{*}{{F-Score}} \\ 
$s_v$       & $s_a$     & $s_p$   &                            &                            &                         &                          \\ \hline
   &       &   \checkmark     &            &  52.40   $\pm$   9.72    &  54.78  $\pm$   8.68     &  50.52   $\pm$  9.24  \\ 
     &     \checkmark   &       &            &  82.48     $\pm$   5.22   &   81.73    $\pm$   4.14      &   80.52   $\pm$  4.95    \\ 
 \checkmark    &       &       &      \checkmark       &     89.83  $\pm$   4.62    &      88.91  $\pm$   4.50      &    88.70  $\pm$   4.69   \\ 
 \checkmark    &      &   \checkmark     &      \checkmark       &     90.24  $\pm$   3.91    &      88.91  $\pm$   3.82     &    88.72  $\pm$   4.83   \\ 
 \checkmark    &   \checkmark      &      &     \checkmark      &     94.23  $\pm$   3.09    &      93.47  $\pm$   3.36      &    93.51  $\pm$   3.29   \\ 
   \checkmark      &   \checkmark      &    \checkmark    &          &     93.55  $\pm$   2.44    &      92.33  $\pm$   2.88      &    92.33  $\pm$  2.90  \\ 
      \checkmark      &   \checkmark      &    \checkmark    &   \checkmark &       94.90  $\pm$   2.23    &      94.34  $\pm$   2.21      &    \textbf{94.34  $\pm$  2.26}   \\ \hline
\end{tabular}
\end{table}

The audio modality is used as complementary data in this study. Nonetheless, in the majority of disappearance cases, no auditory perception is gathered. Furthermore, the sounds that occur when an object hits the table or to the container intermingle in SPC (the anomaly cases where objects are spilled in a pouring task due to overflowing) and SAFE scenarios. As a result, it may not be an essential indicator for these anomaly types.

\begin{figure}[h]
\captionsetup[subfigure]{}
    \centering
    \begin{subfigure}[t]{0.5\textwidth}
        \centering
        \includegraphics[scale=0.6]{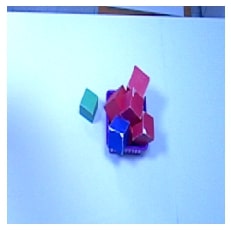}
       \caption{\label{cam-fca-robot}}
    \end{subfigure}%
    ~ 
    \begin{subfigure}[t]{0.5\textwidth}
        \centering
        \includegraphics[scale=0.6]{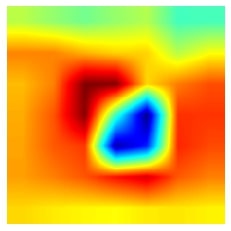}
        \caption{\label{cam-fca-ma}}
    \end{subfigure}%

    \caption{\label{fig:fca-cam} (a) The scene from the RGB-D camera of the robot. (b) CAM of the corresponding FCA case. 
    }
\end{figure}

A class activation map representation of an FCA case is shown in Figure \ref{fig:fca-cam}. Representations based on gradient-weighted class activation mapping (Grad-CAM) \cite{Selvaraju2017grad} are presented in this study. Class activation maps (CAMs) reveal which segments of the input impact the final decision. The contribution of red regions to the final decision is high, while the contribution of blue regions is low. In both figures, the left image represents the images perceived through the robot's camera, while the right image represents the corresponding CAMs. An object falls out of the container on the table in this anomaly type (FCA). The framework appears to be focused on the region where the object that falls out of the container lands rather than the container itself.

\subsubsection{Performance on noisy data}

The performance of the presented CLUE-AI framework is demonstrated on noisy sensory test data in this experiment. Random noise is injected into the perceived data to achieve this (i.e., each pixel of the image is flipped to zero with a given probability value).

\begin{figure}[h]
\begin{center}
     {\includegraphics[scale=0.55]{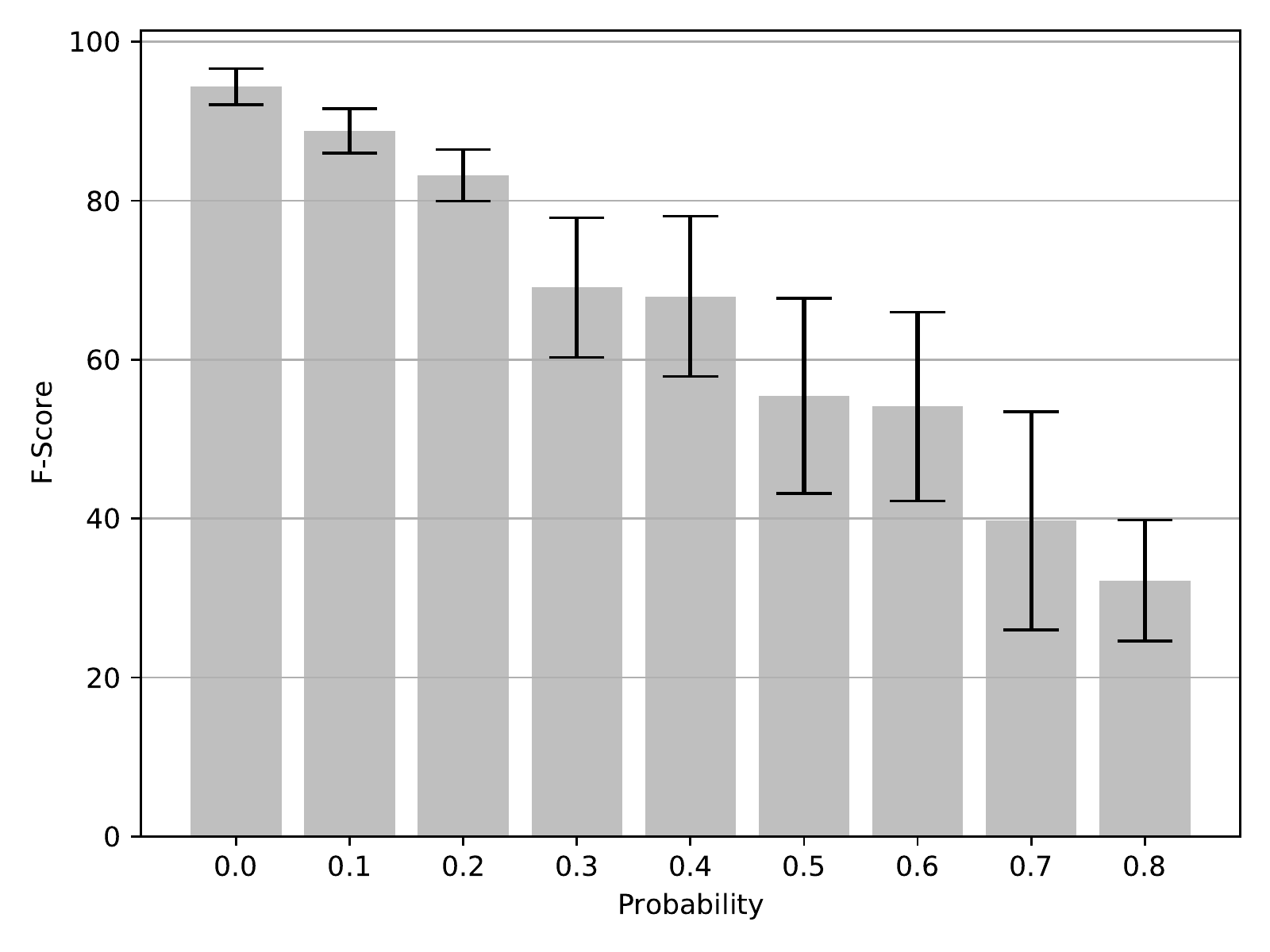}}
\caption{\label{noisy-analysis}Performance analysis of the CLUE-AI framework on noisy data.}
\end{center}
\end{figure}

Figure \ref{noisy-analysis} shows the performance analysis of the CLUE-AI framework for various random noise probabilities with bar graphs. The y-axis corresponds to the f-score value achieved by that setting, and the x-axis corresponds to a distinct noise ratio (i.e., with a probability of that value, the corresponding feature is flipped). The setting with no injected noise provides the highest f-score, as shown by the bars. When the random noise probability is set to 0.3, the performance drops more quickly, and the standard deviation values increase. The framework's performance degrades as the injected noise increases, with an f-score slightly above 30\% for the setting with a noise probability of 0.8.

\subsubsection{Analysis of audio processing}

The CLUE-AI framework treats the auditory stream as 2D data ([time, features]) and convolution layers with 2D kernels are used in the framework. The processing of auditory data with various shaped kernels is evaluated in this study. Rectangular kernels (i.e., (16,4) and (16,5) with strides of the same size) are introduced to the CLUE-AI framework's auditory stage to accomplish this.

\begin{figure}[h]
\begin{center}
     {\includegraphics[scale=0.55]{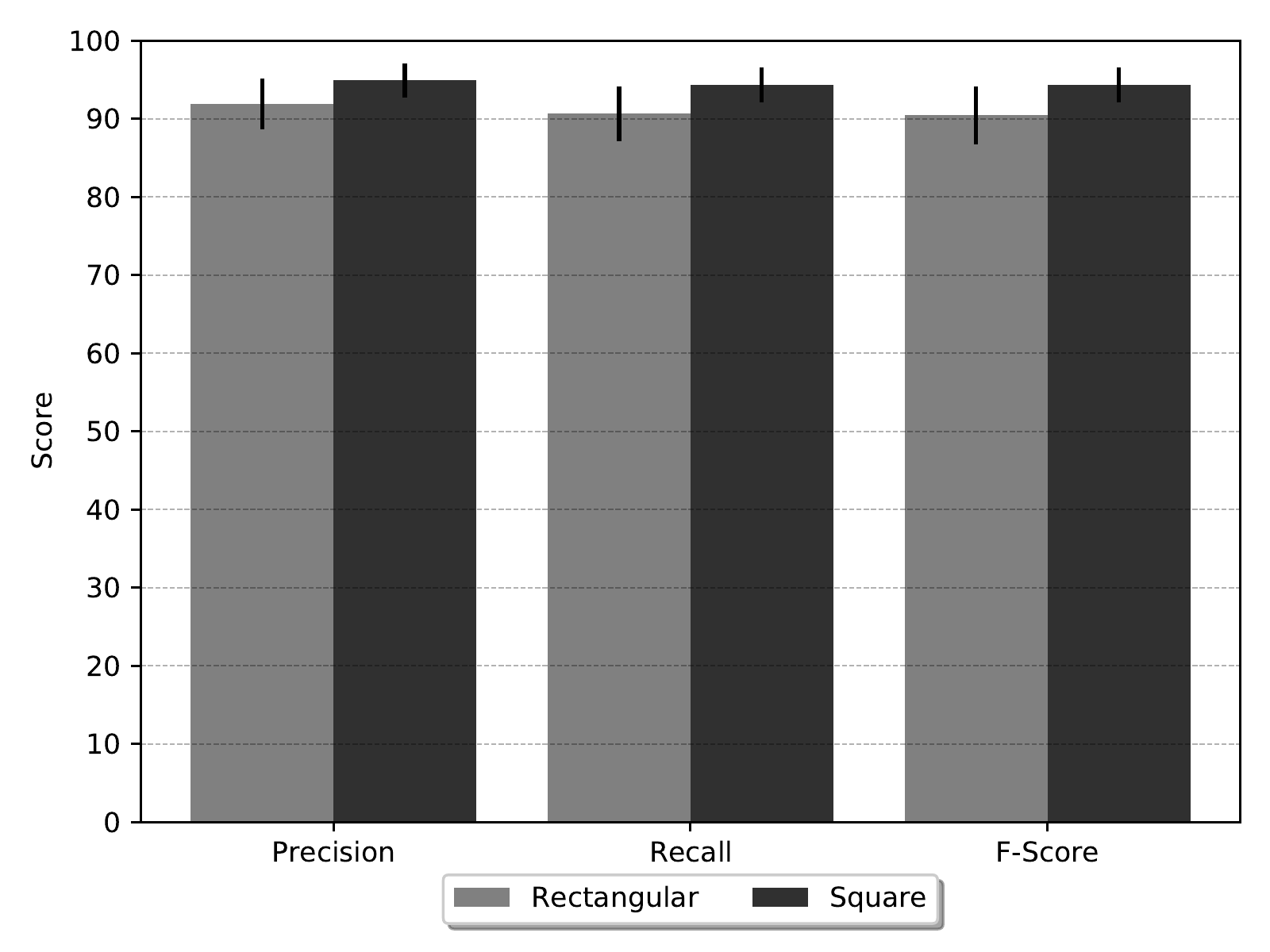}}
\caption{\label{dimension-analysis}Performance analysis of the CLUE-AI framework with different kernels.}
\end{center}
\end{figure}

The scores achieved by different shaped kernels for auditory stream processing are shown in Figure \ref{dimension-analysis}. The x-axis shows the metrics for each setting, while the y-axis shows the scores. For each metric, the first bars (light gray bars) represent the setting with rectangular kernels, while the second bars (dark gray bars) represent the setting with square kernels. As can be seen from the graph, the CNN structure with square kernels on the auditory stream improves each metric, particularly the f-score, which improves by 4\%.

\section{Discussion and Conclusion}

Ethical issues need to be considered for service robots, especially when there are other robots or humans that exist in the environment. The robot should avoid any potential anomaly situations that could lead to failures and harm to itself or others in the environment. This study addresses a critical issue in cognitive robotics by enhancing the self-awareness capabilities of service robots. It presents an essential contribution toward enabling safety awareness procedures in semi-structured environments to perform everyday object manipulation tasks. Continual monitoring of the environment is vital to detect a potential anomaly that could violate safety rules and threaten the environment, robots, or humans. Consequently, the proposed CLUE-AI framework makes use of the necessary diagnostic procedures for enabling robust and safe task execution.

CLUE-AI takes into account three distinct sensory streams, visual, auditory and proprioceptive modalities, to handle anomaly cases that may arise during everyday object manipulation tasks. In the first stream, convolutional visual features are extracted by VGG16 and  fed into self-attention enabled LSTM cells to capture anomaly indicators. In the second one, MFCC features are extracted from the auditory modality, then a CNN block is employed. Last, gripper-related data are processed with a designated CNN layer. These sensory data are combined with a late fusion methodology. The framework is evaluated on real-world scenarios with a Baxter robot performing everyday object manipulation tasks. The feature extraction analysis on the visual stream shows that VGG16 provides better scores in identifying anomalies. Class activation maps are analyzed to investigate the contribution of input images' regions to the identification results. Different kernel shapes are analyzed for processing the auditory stream. The performance on noisy data is presented, and it is shown that the framework is robust to noise up to 30\%. The comparative results indicate that the framework has the ability to identify anomaly cases scoring an f-score of 94\% outperforming the other methods.

Given that the CLUE-AI framework proposed in this study can detect only modeled anomaly cases, the framework could be extended to deal with unknown anomalies that the robot has never encountered before. 

In this study, anomaly identification is modeled as a multi-class classification problem. Our future work includes the representation of anomaly identification as a casual reasoning problem to infer the root causes of potential anomaly cases.

\section{Acknowledgements}
This research was partially supported by Istanbul Technical University Scientific Research Projects (ITU-BAP) under Grant 40240 and the Scientific and Technological Research Council of Turkey (TUBITAK) under Grant 115E-368. The authors would also like to thank Sinan Kalkan for his valuable comments on this research and Arda Inceoglu and Cihan Ak for their contributions to robot experiments. 

\bibliography{identification}

\end{document}